\documentclass[conference]{IEEEtran}

\IEEEoverridecommandlockouts
\usepackage{amsfonts}
\usepackage{amsmath}
\usepackage{graphicx}
\usepackage{color,soul}
\usepackage{subfigure}
\usepackage{multirow}
\usepackage{graphicx}
\usepackage{rotating}
\usepackage{textcomp}
\usepackage{environ}
\usepackage{upquote}
\usepackage{color}
\usepackage{algorithmic}

\usepackage[linesnumbered,ruled,vlined]{algorithm2e}
\SetCommentSty{mycommfont}

\usepackage{array}
\usepackage{amssymb}
\usepackage{caption}
\usepackage{url}
\usepackage{amsmath}
\usepackage{verbatim}

\usepackage[utf8]{inputenc}
\usepackage[english]{babel}
\usepackage{amsthm}
\usepackage{calc}

\ifCLASSINFOpdf
\else
\fi
\begin{document}


\title{\Large \textbf{Prototype Helps Federated Learning: Towards Faster Convergence}}


\author{
\IEEEauthorblockN{Yu Qiao\textsuperscript{1}, Seong-Bae Park\textsuperscript{2}, Sun Moo Kang\textsuperscript{2}, and Choong Seon Hong\textsuperscript{2}*}
\textsuperscript{1}
\textit{Department of Artificial Intelligence, Kyung Hee University, Yongin-si 17104, Republic of Korea}\\
\textsuperscript{2}
\textit{Department of Computer Science and Engineering, Kyung Hee University, Yongin-si 17104, Republic of Korea}\\
Email: \{qiaoyu, sbpark71, etxkang, cshong\}@khu.ac.kr}

\twocolumn[\begin{@twocolumnfalse}      
\maketitle
\begin{abstract}{\normalfont\small\bfseries}      
Federated learning (FL) is a distributed machine learning technique in which multiple clients cooperate to train a shared model without exchanging their raw data. However, heterogeneity of data distribution among clients usually leads to poor model inference. In this paper, a prototype-based federated learning framework is proposed, which can achieve better inference performance with only a few changes to the last global iteration of the typical federated learning process. In the last iteration, the server aggregates the prototypes transmitted from distributed clients and then sends them back to local clients for their respective model inferences. Experiments on two baseline datasets show that our proposal can achieve higher accuracy (at least 1\%) and relatively efficient communication than two popular baselines under different heterogeneous settings.

\vspace{0.5em}
\end{abstract}
\end{@twocolumnfalse}]
\IEEEpeerreviewmaketitle

\section{Introduction}
Federated Learning (FL) was first proposed by Google in 2016 and was originally used to address the update issues of Android terminals. Its original motivation is to carry out privacy-preserving Machine Learning (ML) based on datasets distributed on multiple computing nodes \cite{Google}. Essentially, FL is a distributed ML paradigm. With the further development of Artificial Intelligence (AI), the concept of FL has been further refined and developed. The essential feature of federated training is that the data of all parties are kept locally without revealing privacy issues. In addition, the data distribution of clients in federated scenarios is usually heterogeneous. Therefore, the key challenge is revealed: the data distribution among clients is usually not Independent and Identically Distributed (non-IID), which can lead to poor performance in model inferences~\cite{The non-iid data quagmire, CDFed2023qiao}.

There are existing various studies trying to tackle the non-IID issue. FedAvg \cite{Google} is the first optimization algorithm for federated scenarios. MOON \cite{Model-contrastive} employs a contrastive loss to minimize the difference between the representations learned by global and local models. FedNova \cite{Tackling the objective inconsistency} focuses on the aggregation stage. It lets different clients execute different numbers of local epochs in each global iteration, and then normalizes and scales their local updates before aggregating all local updates to ensure that the aggregated global updates have little biases.

Inspired by \cite{No fear of heterogeneity}, the classifier of the model in the federated settings has a greater bias than other layers. In other words, this means that predictions made through the classifier tend to be highly biased. Therefore, in this paper, we creatively propose to make predictions through the previous layer of the classifier. The output of the previous layer of the classifier is the feature space of the class, and the average of the feature spaces in the same class space can be defined as the "prototype" of this class. Their work \cite{Fedproto} has shown that the information of prototypes can be effectively exploited to resolve heterogeneity in FL.

The main contributions are summarized as follows:
\begin{itemize}
\item We design a novel prototype-based federated learning framework where clients communicate with the server as typical federated training does, but model inferences are made based on the aggregated prototypes instead of the classifier.
\item We propose a prototype aggregation method, in which clients can upload their prototypes to the server for aggregation before finishing the last federation iteration.
\item Experiments over two popular benchmark datasets:
MNIST \cite{MNIST} and Fashion-MNIST \cite{Fashion-MNIST}, show that our proposal has a higher test accuracy (at least 1\% higher) and is relatively efficient in communication than the two popular baselines.
\end{itemize}

\section{Proposed Framework}
\subsection{Problem Statement}
Consider a distributed clients set $\phi$ with private sensitive dataset $\mathcal{D}_i= \{(\boldsymbol {x}_i, y_i)\}$ of size $D_i$ in the distributed edge network. Following the typical training process of FL, clients and the edge server cooperate to train a shared model $\mathcal F(\omega; \boldsymbol {x}_i)$, where $\omega$ is the model parameters of the global model and $\boldsymbol {x}_i$ denotes the feature vector of one client $i$. Our objective is to minimize the loss function across heterogeneous clients as follows:
\begin{equation} \label{global_loss}
    \mathop {\arg\min}_{\omega} {\mathcal{L}} (\omega) = \sum_{i \in \phi} \frac {D_i}{\sum_{i \in \phi}D_i} {\mathcal{L}_i} (\mathcal F(\omega; \boldsymbol {x}_i), y_i),
\end{equation}
where $y_i$ denotes the label of a sample, and $\mathcal{L}_i$ is the empirical risk (e.g. cross-entropy loss) of client $i$, respectively.

\begin{figure}[t]
\centering
\includegraphics[width=0.48\textwidth]{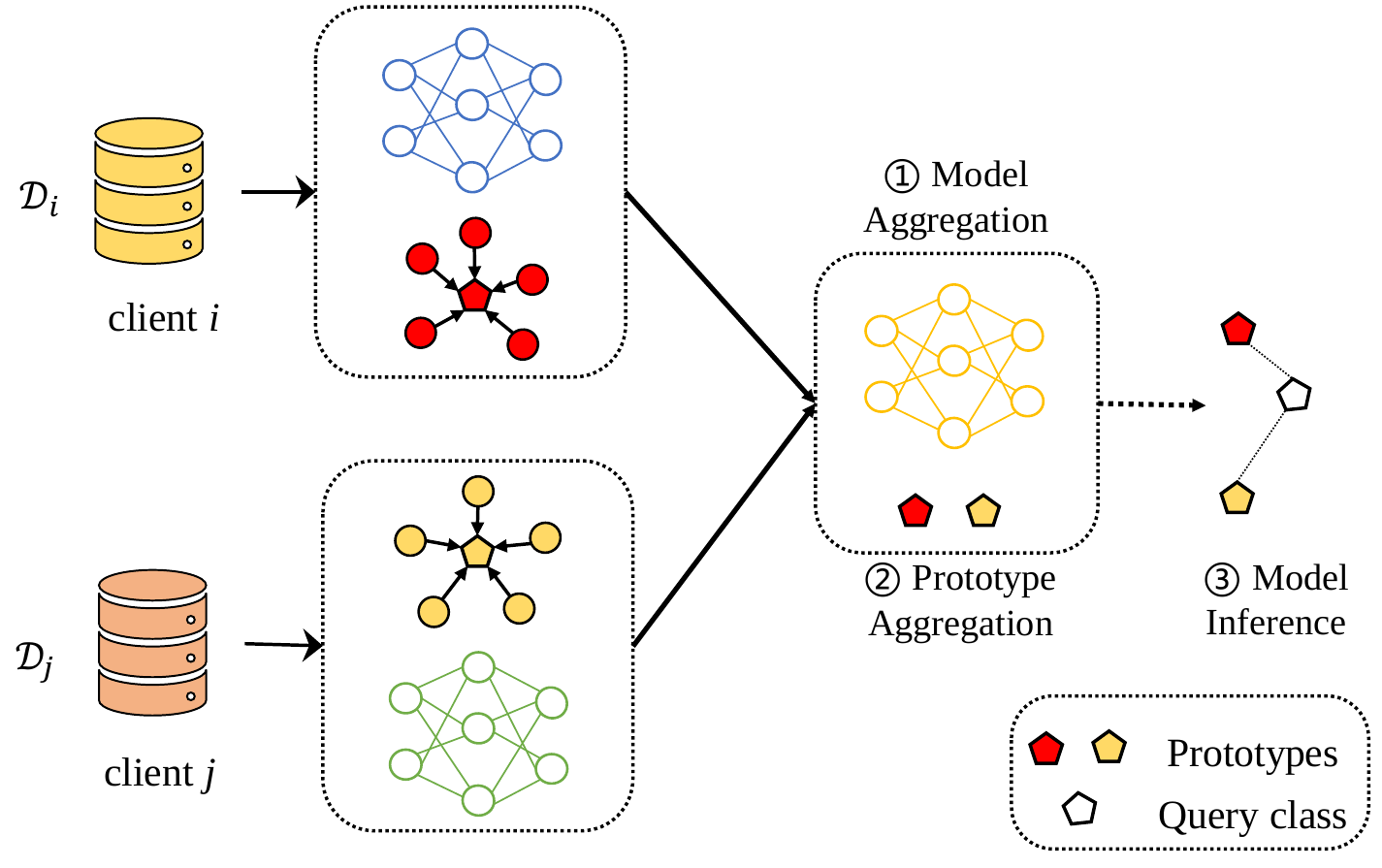}
\caption{The overview of the proposed prototype-based FL framework. Note that in previous $T$-1 global rounds, clients only transmit model parameters with the server to train a shared model, which does not illustrate in the figure. We only illustrate the training process in $T$-th global iteration (i.e. the final global iteration.). In the $T$-th round, clients not only transmit their model parameters, but also their prototypes to the server for model aggregation (step. 1 in this figure) and prototype aggregation (step. 2). Finally, the prediction can be made based on the distance of each query to aggregated prototypes (step. 3).}
\label{System_model}
\end{figure}

\subsection{Proposed Federated Learning Framework}
The typical federated training process can be summarized as: (1). The server distributes the model to each client for local training; (2). Clients update the model parameters individually and send them back to the server for aggregation; (3). The server aggregates these uploaded latest model parameters, and then sends them back again to local clients for the next global iteration, repeating the above process until convergence. 

Our proposal is similar to this typical process, but in the last global round, clients not only need to update their own local model parameters as usual, but also calculate their own prototypes for each class. The calculated prototypes along with model parameters are sent to the server for aggregation. Finally, the aggregated prototypes and the latest model parameters are then sent back to local clients for model inferences. The overview of our proposed framework is presented in Fig. \ref{System_model}.

\subsection{Prototype-based Model Inference Strategy}
Generally, a deep learning model consists of two parts: feature extractor layers and decision-making layers. The former is denoted as  $f_{e}(\omega_{e}; \boldsymbol {x})$, and the latter is denoted as $f_{d}(\omega_{d}; y)$. Therefore, the shared model can be written as $\mathcal F(\omega; \boldsymbol {x})$ =  $f_{d}(f_{e}(\omega_{e}; \boldsymbol {x}), y)$, which means that the output of feature extractor layers acts as the input of the decision-making layers.

\begin{algorithm}[t] 
    \caption{Prototype-based FL (ProtoFed)} 
    \label{alg:Multi-FedProto} 
    \begin{algorithmic}[1]
        \REQUIRE ~~ \\
        Dataset $\mathcal{D}_i$, $\omega_i$
        \STATE \textbf{Model Training:}
        \STATE \textbf{Initialize $\omega^0$}, $\{\overline{y}\}$.
        \FOR{ $t$ = 1, 2, ..., $T$} 
            \FOR{ $i$ = 0, 1,..., $N$ \textbf{in parallel}}
                \IF{$t$ = $T$}
                    \STATE Local model updates.
                    \STATE Prototypes aggregation by Eq. \eqref{proto_agg}.
                \ELSIF {$t$ \textless $T$} 
                     \STATE Local model updates.
                \ENDIF
            \ENDFOR           
        \ENDFOR
        \STATE \textbf{Model Testing:}
        \FOR{ each sample $i$ in testing dataset}
            \FOR{ each class $j$ in $\{\overline{y}\}$}
                \STATE Measure L2 distance between $f_{e}(\omega_{e}; \boldsymbol {x}_i)$
                and $\overline{y}_j$
            \ENDFOR
            \STATE Make final predictions by Eq. \eqref{model_infer}
        \ENDFOR
    \end{algorithmic}
\end{algorithm}

\subsubsection{Prototype Calculation} Following the definition for prototypes given by \cite{Fedproto}, the prototype representing $j$-th class of client $i$-th can be formulated as:
\begin{equation} \label{proto_define}
\widetilde{y}_{i,j} = \frac{1}{D_{i,j}} \sum_{(x,y)\in \mathcal{D}_{i,j}}f_{e}(\omega_{e}; \boldsymbol {x}),
\end{equation}
where $\mathcal{D}_{i,j}$ is the distribution of client $i$-th belong to the $j$-th class, and $D_{i,j}$ is the size for $\mathcal{D}_{i,j}$.

\subsubsection{Global Prototype Aggregation} As calculated by Eq. \ref{proto_define}, these computed prototypes can be sent to the server for aggregation, which can be defined as follows:
\begin{equation} \label{proto_agg}
\overline{y}_j =  \frac{1}{|\phi|} \sum_{i\in \phi} \widetilde{y}_{i,j},
\end{equation}
where $|\phi|$ is the number of clients in the network. This Equation means to average the prototypes from those clients with $j$-th class as global prototypes.

\subsubsection{Prototype-based Model Inference} The results of Eq.\ref{proto_agg} can be viewed as a global representation of each class for model inference. Therefore, the prediction can be made by measuring the L2 distance between the local prototype $f_{e}(\omega_{e}; \boldsymbol {x})$ and the global representation prototypes $\overline{y}_j$, which can be expressed as \cite{qiaoyu2023framework}:
\begin{equation} \label{model_infer}
    \hat y = \mathop {\arg\min}_{j} \|f_{e}(\omega_{e}; \boldsymbol {x})  -  \overline{y}_j\|_2,
\end{equation}
where the final predicted label is denoted as $\hat y$. Further details about prototype-based federated learning are shown in Algorithm.1.

\begin{figure}[t]
\centering
\subfigure[$\alpha$ = 0.05]{\includegraphics[height=4cm,width=4.25cm]{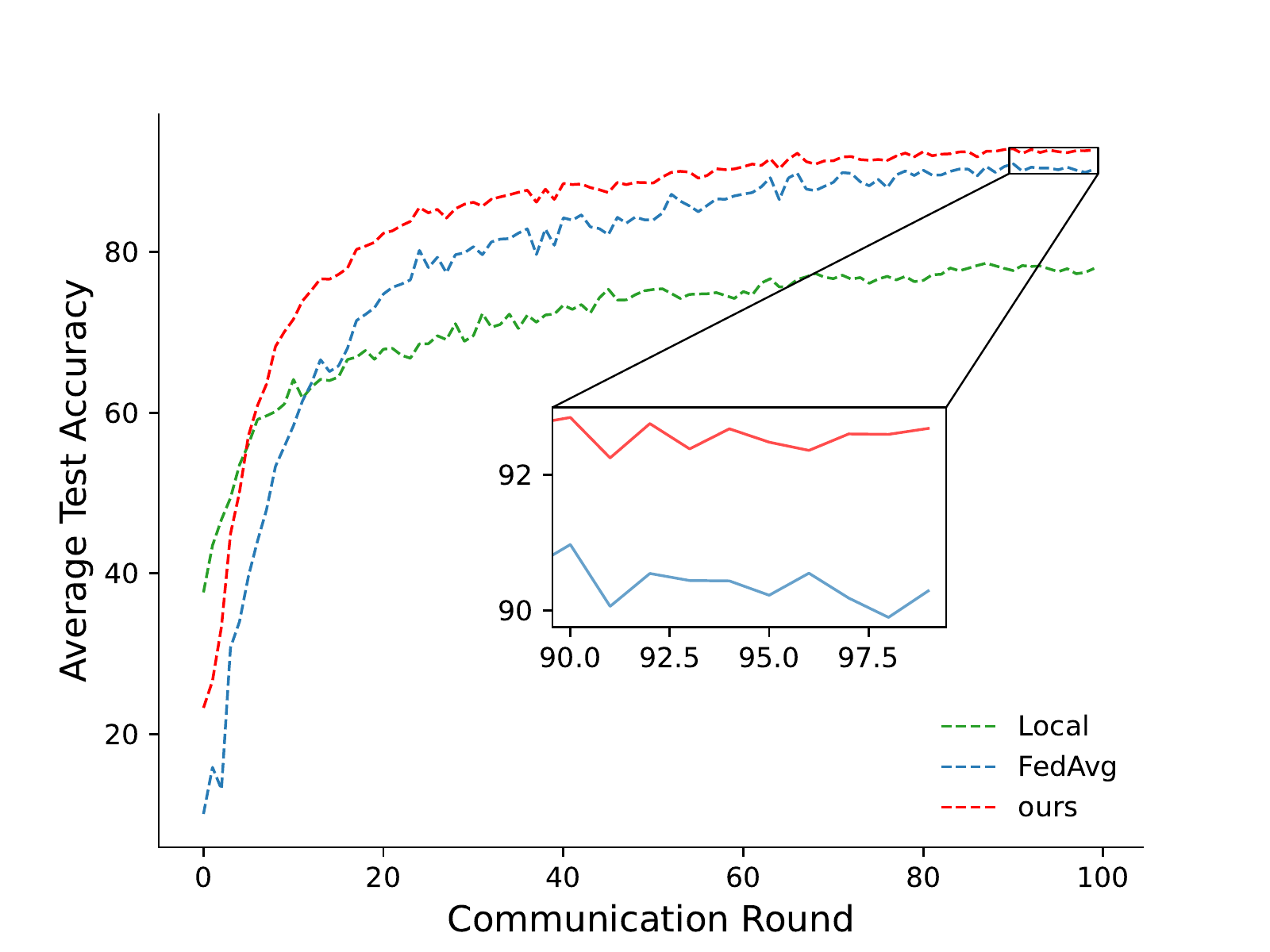}}
\subfigure[$\alpha$ = 0.1]{\includegraphics[height=4cm,width=4.25cm]{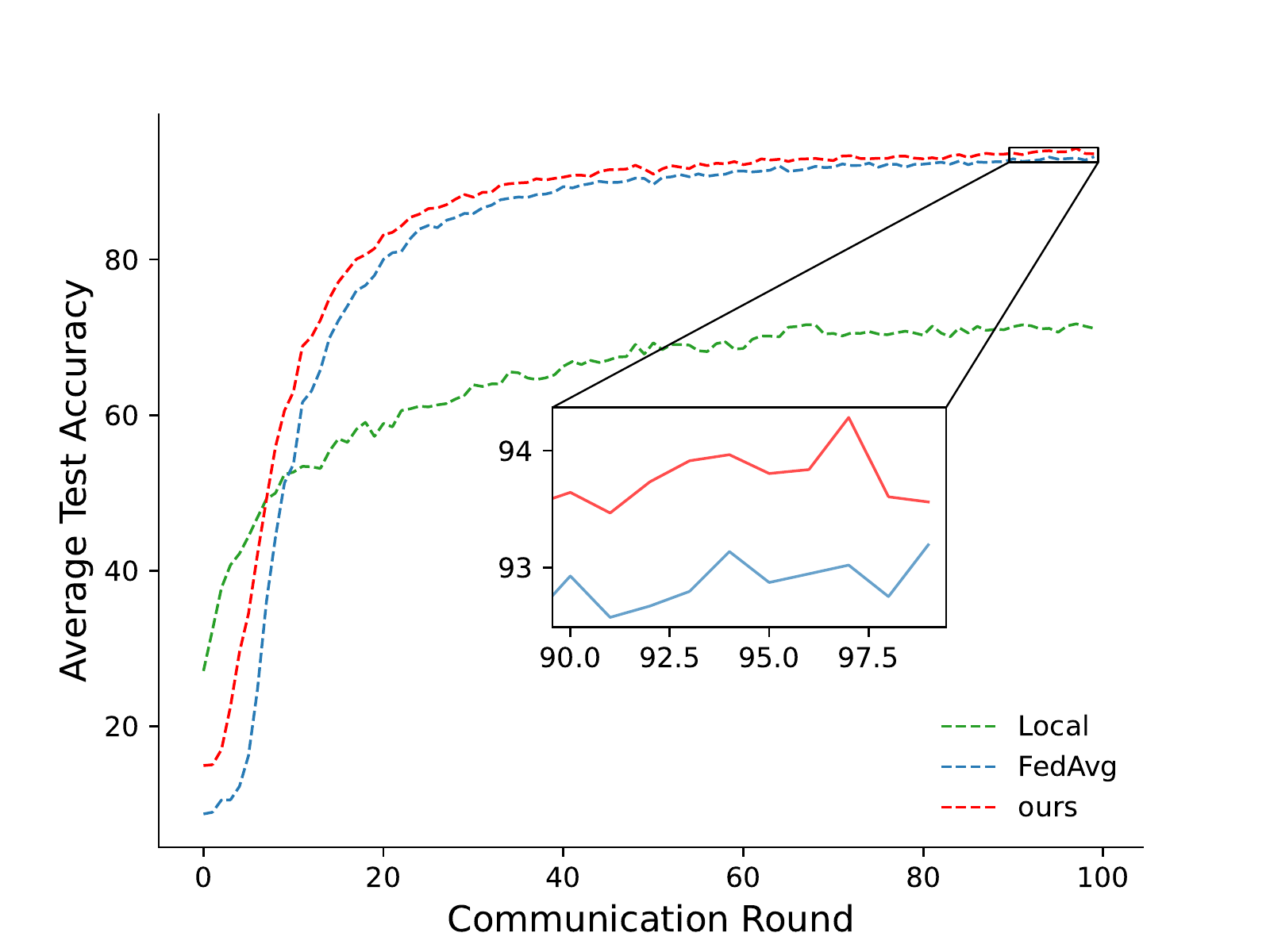}}
\caption{The top-1 average test accuracy of all methods on MNIST in different communication rounds with the degree of skewness $\alpha$ = 0.05 and $\alpha$ = 0.1.}
\label{mnist_0.05_0.1}
\end{figure}

\begin{figure}[t]
\centering
\subfigure[$\alpha$ = 0.05]{\includegraphics[height=4cm,width=4.25cm]{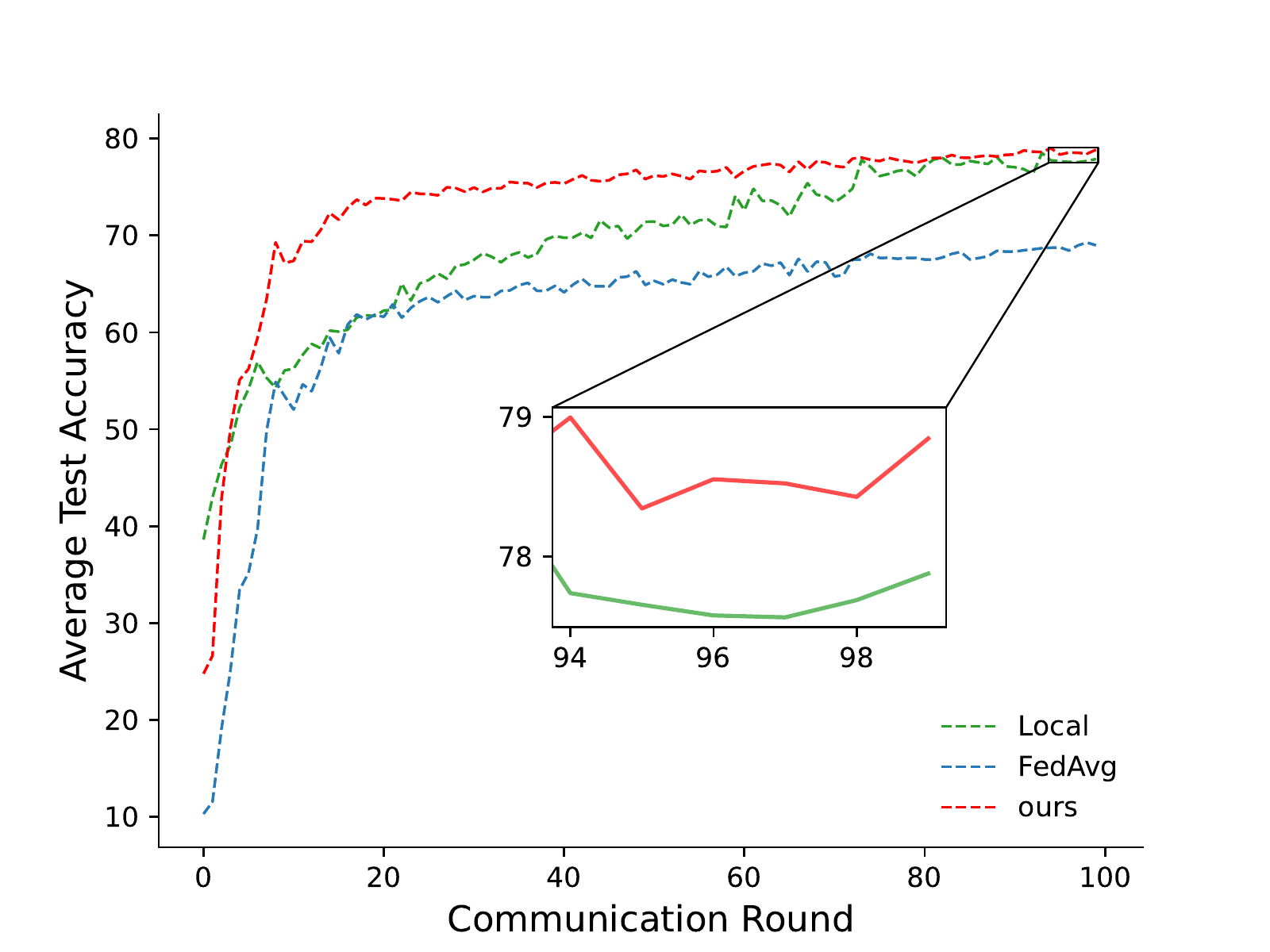}}
\subfigure[$\alpha$ = 0.1]{\includegraphics[height=4cm,width=4.25cm]{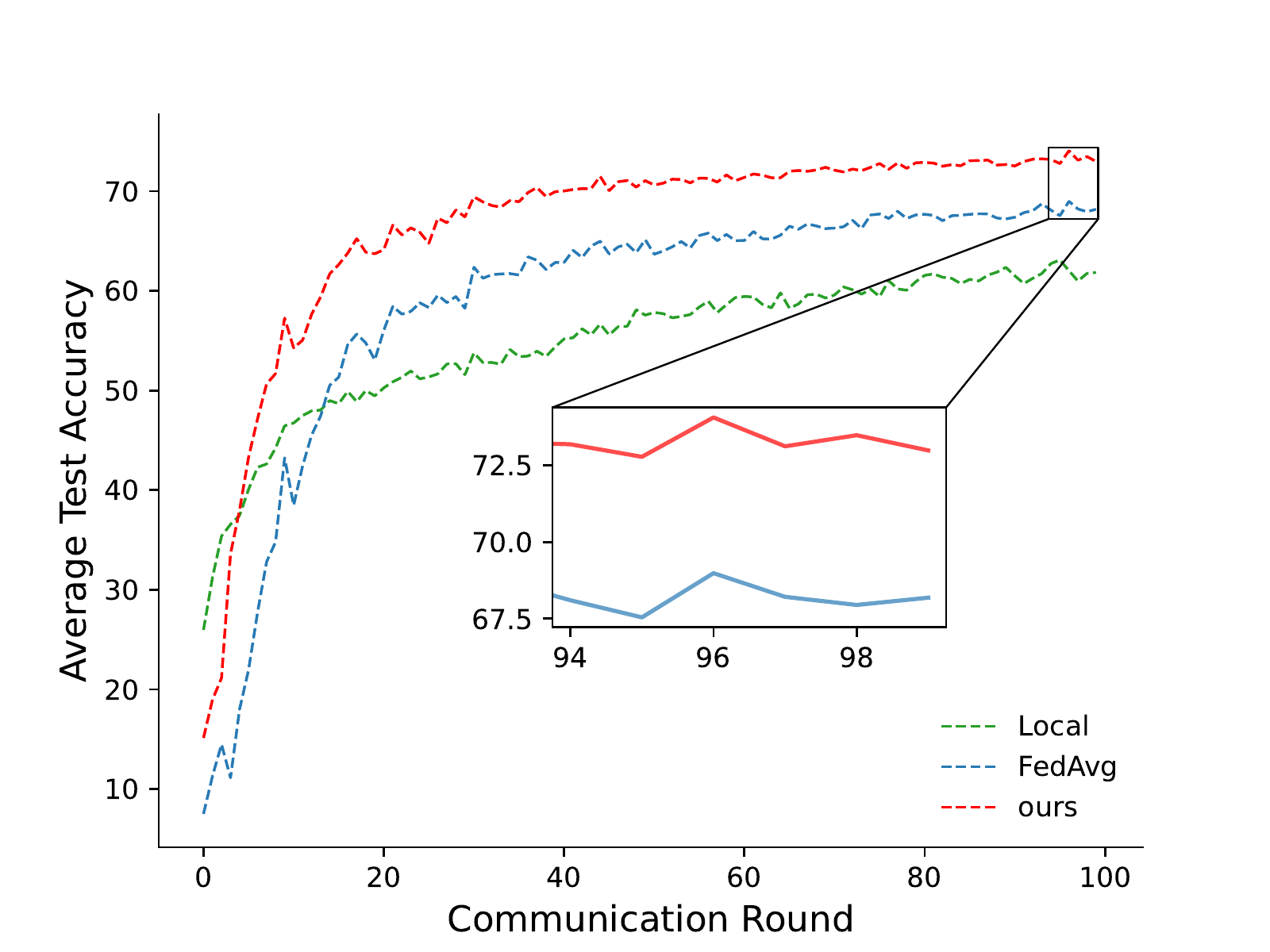}}
\caption{The top-1 average test accuracy of all methods on Fashion-MNIST in different communication rounds with the degree of skewness $\alpha$ = 0.05 and $\alpha$ = 0.1.}
\label{fmnist_0.05_0.1}
\end{figure}

\section{Experiments}
\subsection{Dataset and Local Model}
We use MNIST \cite{MNIST} and Fashion-MNIST \cite{Fashion-MNIST} as benchmarks for comparison. MNIST is a handwritten digit recognition dataset with 10 digits and an image size of 28x28x1. Fashion-MNIST is more complex than MNIST, covering a total of 70,000 images of different items in 10 categories and an image size of 28x28x1. We use a 4-layer CNN network with 2 convolutional layers and 2 fully connected layers for MNIST and Fashion-MNIST, similar networks are also adopted in their work \cite{No fear of heterogeneity, Fedproto}.

\subsection{Implementation Details}
We use 20 clients for all experiments, and the default training parameters for the MNIST and Fashion-MNIST datasets are set to $B$ = 8, $E$ = 1, $\eta$ = 0.01, representing the local batch size, local epochs, and learning rate, respectively. We sample 5000 samples from the training dataset and distribute them to all clients to mimic the situation where training samples are limited in the real-world \cite{Fedproto}. Further, we adopt the Dirichlet distribution \cite{Dirich} to simulate the heterogeneous setting among clients, which can be expressed as Dir($\alpha$), where the smaller $\alpha$, the more unbalanced the data distribution among clients is.

\subsection{Accuracy and Communication Efficiency Comparison}
We compare our proposal with two popular baselines \texttt{Local} and \texttt{FedAvg} under different heterogeneities, where \texttt{Local} indicates that clients train their own model individually without any communication with the server or other clients, and \texttt{FedAvg} as the first federated algorithm is the most popular baseline. Note that we perform prototype-based model inference in each round during training in order to compare their test accuracies after each global iteration.

The accuracy and communication efficiency for comparison of all methods on MNIST and Fashion-MNIST are shown in Fig.\ref{mnist_0.05_0.1} and Fig.\ref{fmnist_0.05_0.1}, respectively. It appears that our prototype-based model inference strategy achieves higher test accuracy and a relatively faster convergence rate in each global communication round than other methods under different heterogeneous settings on both these two datasets.

To be more specific, our proposal on MNIST outperforms \texttt{Local} and \texttt{FedAvg} by 18.9\% and 2.4\% in terms of accuracy when $\alpha$ = 0.05, and also outperforms them by 31.5\% and 1.0\%  when $\alpha$ = 0.05, respectively. Further, our proposal on Fashion-MNIST outperforms \texttt{Local} and \texttt{FedAvg} by 1.4\% and 14.3\% in terms of accuracy when $\alpha$ = 0.05, and also outperforms them by 18.4\% and 7.4\% when $\alpha$ = 0.1, respectively.

\section{Conclusion}
In this paper, a prototype-based federated learning framework is proposed, which can help boost the performance of model inference. We first present the prototype calculation method, and then we introduce the prototype aggregation approach. Finally, we propose the prototype-based model inference strategy. Experiments show that our strategy can improve the accuracy by at least 1\% on MNIST and Fashion-MNIST compared to two popular baselines \texttt{Local} and \texttt{FedAvg}, and can achieve relatively efficient communication. For future works, our proposal has the potential to be combined with other state-of-the-art methods and tests on more datasets.


\begin{thebibliography}{1}
\bibitem{Google} McMahan B, Moore E, Ramage D, et al. ``Communication-efficient learning of deep networks from decentralized data,`` {\em Artificial intelligence and statistics}, pp. 1273-1282, 2017.
\bibitem{The non-iid data quagmire} Kevin Hsieh, Amar Phanishayee, Onur Mutlu, and Phillip Gibbons. ``The non-iid data quagmire of decentralized machine learning,`` {\em International Conference on Machine Learning}, pp. 4387–4398, 2020.
\bibitem{Model-contrastive} Li, Q., B. He, D. Song. ``Model-contrastive federated learning,`` {\em IEEE/CVF Conference on Computer Vision and Pattern Recognition}, 2021.
\bibitem{Tackling the objective inconsistency} J. Wang, Q. Liu, H. Liang, G. Joshi, and H. V. Poor. ``Tackling the objective inconsistency problem in heterogeneous federated optimization,`` {\em Advances in Neural Information Processing Systems}, vol. 33, 2020.
\bibitem{No fear of heterogeneity} Luo M, Chen F, Hu D, et al. ``No fear of heterogeneity: Classifier calibration for federated learning with non-iid data,``  {\em Advances in Neural Information Processing Systems}, vol. 34, pp. 5972-5984, 2021.
\bibitem{Fedproto} Tan Y, Long G, Liu L, et al. ``Fedproto: Federated prototype learning across heterogeneous clients,``  {\em AAAI Conference on Artificial Intelligence}, vol. 1. no. 3, 2022.
\bibitem{MNIST} LeCun, Yann, et al. ``Gradient-based learning applied to document recognition,`` Proceedings of the IEEE, vol. 86, no.11, pp. 2278-2324, 1998.
\bibitem{Fashion-MNIST} Xiao, Han, Kashif Rasul, and Roland Vollgraf. ``Fashion-mnist: a novel image dataset for benchmarking machine learning algorithms,`` arXiv preprint arXiv:1708.07747, 2017.
\bibitem{Dirich} Yurochkin M, Agarwal M, Ghosh S, et al. ``Bayesian nonparametric federated learning of neural networks," {\em International Conference on Machine Learning}, pp: 7252-7261, 2019.
\bibitem{qiaoyu2023framework}{Qiao, Yu, et al. "A Framework for Multi-Prototype Based Federated Learning: Towards the Edge Intelligence." 2023 International Conference on Information Networking. IEEE, 2023.}
\bibitem{CDFed2023qiao}{Qiao, Yu and Munir, Md Shirajum, et al. "CDFed: Contribution-based Dynamic Federated Learning for Managing System and Statistical Heterogeneity." IEEE/IFIP Network Operations and Management Symposium, 2023.}

\end{thebibliography}
\end{document}